\documentclass[11pt,a4paper]{article}
\usepackage[T1]{fontenc}
\usepackage{acl2021}
\usepackage{times}
\usepackage{latexsym}
\usepackage{amsmath}
\usepackage{mathrsfs}
\usepackage{bbm}
\usepackage{amsfonts}
\usepackage{mathtools}
\usepackage{algorithm}
\usepackage{placeins}
\usepackage{float} 
\usepackage{varioref}
\usepackage{amssymb}

\usepackage{mwe} 
\usepackage{caption}

\usepackage{algorithmicx}
\usepackage{algpseudocode}
\algnewcommand\algorithmicinput{\textbf{\small{INPUT:}}}
\algnewcommand\INPUT{\item[\algorithmicinput]}

\algnewcommand\algorithmicoutput{\textbf{OUTPUT:}}
\algnewcommand\OUTPUT{\item[\algorithmicoutput]}
\usepackage{changes}

\algnewcommand\algorithmicbuild{\textbf{Compute layers embeddings:}}
\algnewcommand\e{\item[\algorithmicbuild]}

\usepackage{cancel}
\setdeletedmarkup{\cancel{#1}}

\definechangesauthor[color=green]{PC}

\definechangesauthor[color=orange]{GS}

\usepackage{microtype}
\usepackage{subfig}
\usepackage{booktabs}

\title{Automatic Text Evaluation through the Lens of Wasserstein Barycenters}

\author{\textbf{Pierre Colombo\textsuperscript{1,2}, Guillaume  Staerman\textsuperscript{1}}, \textbf{Chloe Clavel\textsuperscript{\rm 1},  Pablo Piantanida\textsuperscript{$^*$}} \\
\textsuperscript{\rm 1}LTCI, Telecom Paris, Institut Polytechnique de Paris,\\ \textsuperscript{\rm 2}IBM GBS France,\\
$^*$CentraleSupelec-CNRS, Universite Paris-Saclay \\
\textsuperscript{\rm 1}firstname.lastname@telecom-paris.fr, \\
\textsuperscript{\rm 2}pierre.colombo@ibm.com, \\
\textsuperscript{\rm $^*$}pablo.piantanida@centralesupelec.fr,
}

\date{}

\begin{document}
\maketitle
\begin{abstract}
A new metric \texttt{BaryScore} to evaluate text generation based on deep contextualized embeddings (\textit{e.g.}, BERT, Roberta, ELMo) is introduced. This metric is motivated by a new framework relying on optimal transport tools, \textit{i.e.}, Wasserstein  distance and barycenter. By modelling the layer output of deep contextualized embeddings as a probability distribution rather than by a vector embedding; this framework provides a natural way to aggregate the different outputs through the Wasserstein space topology. In addition, it provides theoretical grounds  to our metric and offers an alternative to available solutions (\textit{e.g.},  MoverScore and BertScore). Numerical evaluation is performed on four different tasks:  machine translation, summarization, data2text generation and image captioning. Our results show that \texttt{BaryScore} outperforms other BERT based metrics and exhibits more consistent behaviour in particular for text summarization.  
\end{abstract}

\section{Introduction}
Automatic Evaluation (AE) of Natural Language Generation (NLG) is a key problem towards better systems \cite{specia2010machine}. It allows to assess the quality of generated text without relying on human evaluation campaigns that are expensive and time consuming \cite{belz2006comparing,sai2020survey}. For instance, it becomes crucial to design automatic and effective metrics with simultaneous goals: (i) to be able to compare, to control and to debug systems without relying on human annotators \cite{peyrard-2019-simple,peyrard2019studying}; and (ii) to improve the learning phase of models by deriving losses that are better surrogate of human judgment than the widely used cross-entropy loss  \cite{clark-etal-2019-sentence}.


A plethora of automatic metrics has been introduced these last few years and may be grouped into two general classes: trained \cite{ma2017blend,shimanaka2018ruse,lowe2016evaluation,lita2005blanc} and untrained metrics \cite{doddington2002automatic,popovic2015chrf}. In this paper, we mainly focus on untrained metrics that can be further split into three subgroups: \textit{string matching} \cite{bleu,lin-2004-rouge,banerjee2005meteor,doddington2002automatic,popovic2015chrf}, \textit{edit based} \cite{leusch-etal-2006-cder,snover2006ter,wang2016character} and \textit{embedding based} metrics \cite{chow-etal-2019-wmdo,kusner2015wmd,lo2011meant,lo2019yisi}. Both string matching and edit based metrics fail to assign reliable scores when reference and candidate convey the same meaning with distinct surface forms \cite{reiter2009investigation} (\textit{e.g.}, case of synonyms and paraphrases). These shortcomings have been addressed by metrics based on continuous representations. Recently, they have benefited from contextual embeddings such as ELMO \cite{elmo} and BERT \cite{bert,roberta}. Perhaps, the most known are BERTScore \cite{zhang2019bertscore}, MoverScore \cite{zhao2019moverscore} or Sentence Mover \cite{clark-etal-2019-sentence} which optimise Word Mover Distance (WMD) \cite{kusner2015wmd}\footnote{\cite{zhao2019moverscore} shows that BertScore can be viewed as a non-optimized transport problem.}, a particular instance of optimal transport (OT) problem. 
\\ Originally introduced by \citet{kusner2015wmd} the WMD is used to compute the Wasserstein distance between text documents relying on a \textit{single layer} embedding such as GloVe \cite{pennington2014glove} or Wor2Vect \cite{word2vec}. To apply WMD with \textit{multi layers} embedding several recipes have been proposed. BertScore selects the best layer based on a validation set. However, the selection of the validation set is arbitrary while on the other hand, a single layer selection does not exploit the information available in other layers \cite{voita2019bottom,hewitt2019designing,liu2019linguistic}. MoverScore and Sentence Mover attempt to leverage the information available in other layers by aggregating the layers using a power mean \cite{ruckle2018concatenated}. In addition to adding extra hyper-parameters, this aggregation method relies on euclidean topology which induces a geometrical discrepancy as the final cost is computed using a Wasserstein distance. 

\noindent \textbf{Our contributions.} We introduce \texttt{BaryScore} a novel metric, which addresses aforementioned aggregation pitfalls by relying on Wasserstein barycenters and evaluate its performance on four different tasks: neural machine translation, text summarization, image captioning and data2text generation. Our main contributions can be summarized as follow:

\textbf{1.} \textit{A novel metric to measure the semantic equivalence between two texts}. This metric  relies on the  embedding geometry of the layers induced in Wasserstein spaces. In order to overcome the geometric distortion generated by the aggregation techniques used to compute WMD with deep embeddings (\textit{e.g.}, BERT, ELMo and Roberta), we aggregate layers information using the Wasserstein barycenter. This new formulation offers a topological advantage, \textit{i.e.}, using barycenters giving meaning to the use of OT based distance afterwards and is parameter-free, i.e., it avoids choosing by hand the best layer (as for BertScore) or selecting the exponent in the power means (as for  MoverScore). Our formulation provides an alternative and a generalization to the WMD formulation \cite{kusner2015wmd} (originally introduced for Word2Vec) when applied to embeddings which are coming from multi-layer neural networks and thus, it provides theoretical motivation to \texttt{BaryScore}\footnote{Bary stands for Barycenter.} a new metric that aggregates deep contextualized embedding using Wasserstein barycenters. 

\textbf{2.} \textit{Applications and numerical results.} We demonstrate that \texttt{BaryScore} provides better results than a large variety of state-of-art untrained metrics on four text generation tasks: namely NMT, summarization, image captioning, data2text suggesting that Wasserstein barycenters offer a promising direction moving forward.

\section{Related Work}
The goal of NLG is to generate coherent, readable and informative text from some input data (\textit{e.g.}, texts, images and tables). However, the exact definition of each of these three criteria remains task-dependent and thus, making it hard to provide a unique metric for all tasks. As an example, NMT focuses on fluency, fidelity and adaquatie \cite{hovy1999toward,white1994arpa} in contrast to summarization where annotators have to focus on coherence, content, readability, grammatically, coherence and conciseness \cite{mani2001automatic}. In the following, we describe for each of the  four considered tasks (\textit{i.e.}, NMT, text summarization, image captioning and data2text generation) the most used metrics. 

\noindent \textbf{Metrics for NMT.} Most of the metrics commonly used in NMT rely on comparing surface form (\textit{e.g.}, word, subword, n-gram overlap and edit based distances \cite{levenshtein1966binary}) between text and candidates. Perhaps the most popular metrics  are the ones used for WMT shared tasks \cite{mathur2020results,ma-etal-2019-results,ma2018results,bojar-etal-2017-results} which include SENTBLEU, BLEU \cite{bleu}, CHARACTER \cite{wang2016character}, COMET \cite{rei-etal-2020-comet}, YISI \cite{lo-etal-2018-accurate}, MEE \cite{mukherjee2020mee}, EED \cite{stanchev-etal-2019-eed}, CHRF \cite{popovic2015chrf,popovic2017chrf++}, ESIM \cite{chen2016enhanced}, PRISM \cite{thompson2020automatic} to only mention a few among others. A new family of metrics based on pretrained transformers (\textit{i.e.}, BertScore, MoverScore) has recently emerged with very good performance in NMT, incorporating deeper semantic information through contextualized representations. 

\noindent\textbf{Metrics for summarization.} Designing better summarization metrics is an active area of research \cite{scialom2021safeval} and many of these metrics can be further optimized to produce better summaries \cite{bohm2019better}. Popular metrics include machine translation metrics (\textit{i.e.}, CHRF, BLEU, METEOR \cite{banerjee2005meteor,guo-hu-2019-meteor,denkowski2014meteor}, BertScore, MoverScore or SentenceMover \cite{clark-etal-2019-sentence}), ROUGE \cite{lin-2004-rouge,ganesan2018rouge} or data statistics (\textit{e.g.},  density and  compression ratio) \cite{grusky2018newsroom}. 

\noindent \textbf{Metrics for data2text.} Data2text generation aims at generating text from structured data \cite{kim2010generative,chen2008learning,wiseman2017challenges}. In the present work, we focus on the WebNLG 2020 challenge \cite{perez2016building,gardent2017creating} which ranks the system using five automatic metrics: BLEU, METEOR, BERTScore, TER and CHRF++.

\noindent \textbf{Metrics for image captioning.} Task specific metrics for image captioning include CIDEr  \cite{vedantam2015cider} that rely on n-grams, LEIC \cite{cui2018learning}  using scene graph similarity and pretrained metrics such as SPICE \cite{anderson2016spice}. In recent work by \cite{zhang2019bertscore,zhao2019moverscore}, these metrics are compared with NMT specific metrics (e.g., BLEU and METEOR).

\section{Background on Optimal Transport}
The Wasserstein distance (\textit{i.e.}, Earth Mover Distance) which arises from the idea of optimal transport provides a way to measure dissimilarities between two probability distributions. Due to its appealing geometric properties, it has found many applications in machine learning such as generative models \cite{arjovsky2017,WAE,gulrajani2017improved}, domain adaptation \cite{courtyAD}, clustering \cite{clustering_wass,ye2017}, adversarial examples \cite{wong_adversarial}, robustness \cite{staerman2021} or NLP \cite{kusner2015wmd,zhao2019moverscore,singh2020context}.
First designed as an optimal transport optimization problem, it relies on minimizing a transport cost between points drawn from all possible coupling measures. Its ability to take into account the underlying geometry of the space as well as capture information from distributions with non-overlapping supports makes it a powerful alternative to several dissimilarity measures such as the family of $f$-Divergences.

\noindent \textbf{Wasserstein distance.} Let $\mathcal{M}_{+}^1(\mathbb{R}^d)$ denote the space of all probability distributions defined on $\mathbb{R}^d$ with $d\in \mathbb{N}^*$. The Wasserstein distance between two arbitrary measures $\mu\in \mathcal{M}_+^{1}(\mathcal{X})$ and $\nu \in \mathcal{M}_+^{1}(\mathcal{Y})$ is defined through the resolution of the Monge-Kantorovitch mass transportation problem \citep{Villani,Peyre}:
\begin{equation}\label{OT-primal}
\hspace{-0.1em}\mathcal{W}(\mu, \nu) \hspace{-0.1em}= \hspace{-1em}\underset{ \pi~\in~\mathcal{U}(\mu,\nu)}{\min} \hspace{-0.2em}  \left(\int_{\mathcal{X} \times \mathcal{Y}}  \hspace{-0.5cm} ||x-y||^p d\pi(x, y) \right)^{1/p} \hspace{-1.4em} ,
\end{equation}
where $\mathcal{U}(\mu,\nu)= \{ \pi \in \mathcal{M}^1_+(\mathcal{X} \times \mathcal{Y}): \; \; \int \pi(x,y)dy =\mu(x) ;\int \pi(x,y) dx=\nu(y) \}$  is the set of joint probability distributions with marginals $\mu$ and $\nu$.  In the remainder of this paper, we focus on the Wasserstein distance associated with the quadratic cost, \textit{i.e.,} $p=2$. Thus, the Wasserstein distance aims to find the best possible way to transfer the probability mass from $\mu$ to $\nu$ while minimizing the transportation cost defined by the euclidian distance.





\begin{figure}
\centering
\begin{tabular}{cc}
{\small Euclidian} &{\small Wasserstein}\\
   \includegraphics[scale=0.25, trim= 1cm 0cm 0cm 0cm]{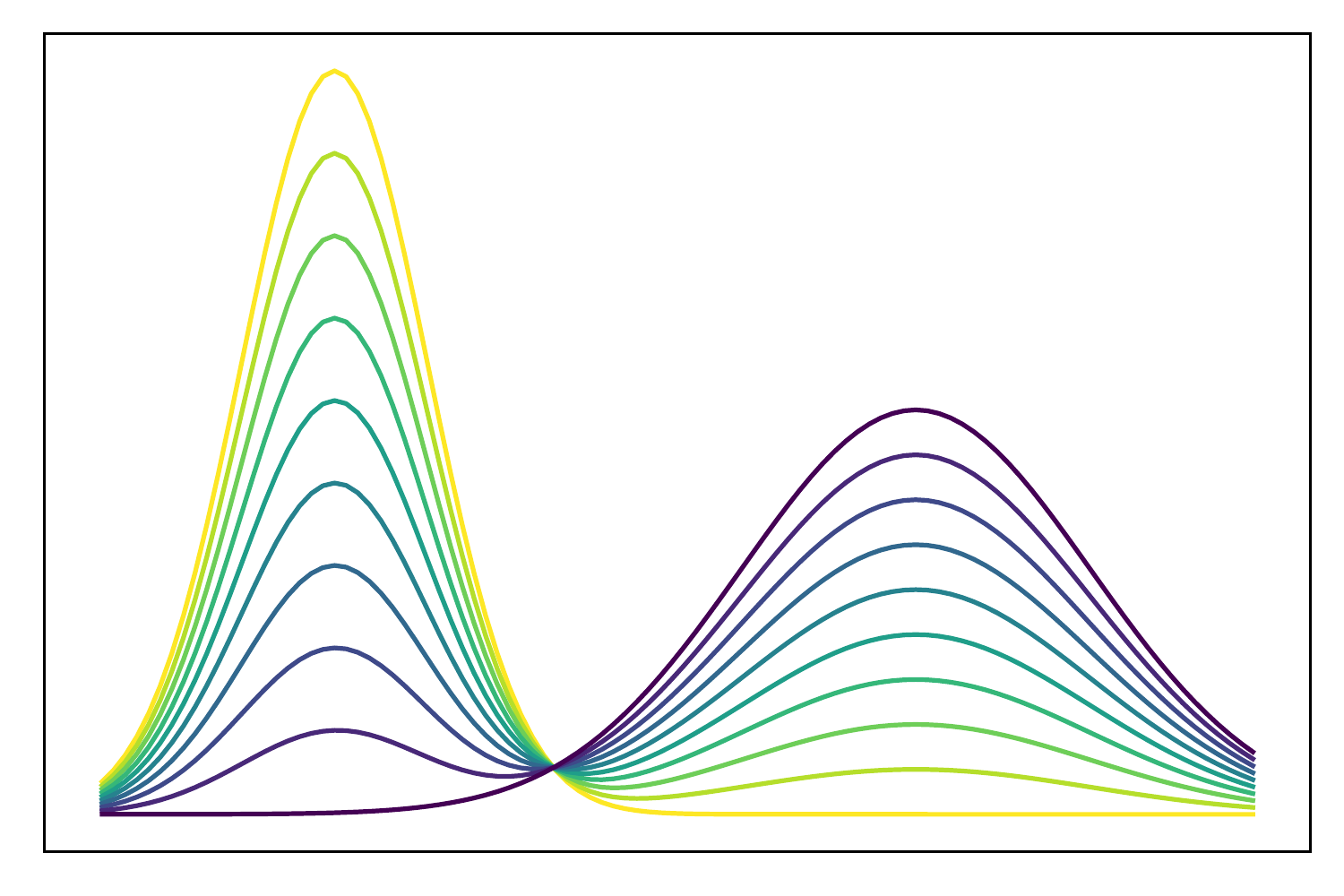}  &  \includegraphics[scale=0.25, trim= 1cm 0cm 0cm 0cm]{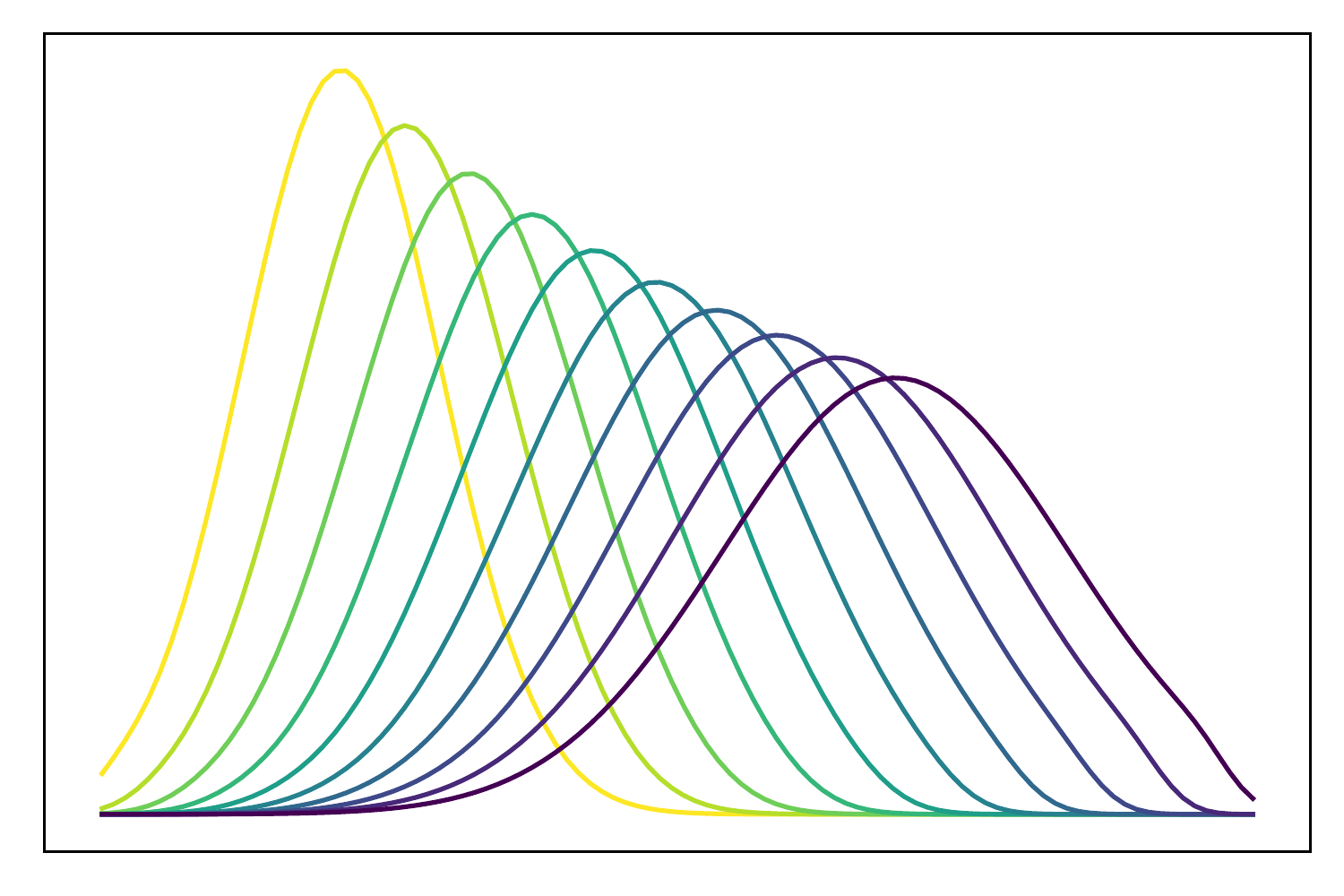}
\end{tabular}
    \caption{Euclidian (left) and Wasserstein (right) interpolation between densities of two Gaussian distributions.}
    \label{fig:barycenter}
\end{figure}

\noindent \textbf{Wasserstein barycenters.} Because optimal transport is based on mass displacement, it also defines an interesting way to interpolate between several input measures. The Wasserstein barycenter, first introduced and studied in \cite{Carlier}, defines an interpolation measure between several probability distributions. The main asset of Wasserstein barycenters is  to take into account the geometry of the space where input measures live in (cf. Fig.~\ref{fig:barycenter}). Given $N$ probability distributions: $\mu_1,\ldots, \mu_N \in \mathcal{M}_+^{1}(\mathbb{R}^d)$ and weights $(\alpha_1,\ldots, \alpha_N) \in \mathbb{R}_+$, the Wasserstein barycenter optimization problem of these distributions w.r.t. the weights  is defined as:
\begin{equation}\label{OT-barycenter}
\mu = \underset{ \mu \in \mathcal{M}_{+}^{1}(\mathbb{R}^d)}{\text{argmin}} \;  \sum_{i=1}^{N} \alpha_i \mathcal{W}(\mu_i, \mu),
\end{equation}
where the support of $\mu$ may be unknown. \autoref{OT-barycenter} defines a weighted average in the Wasserstein space. To make it computationally tractable, the measure $\mu$ is often constrained to be a discrete measure with free \cite{cuturi_barycenter,alvarezesteban2016,cuturi2016,giulia2019} or fixed support \cite{benamou2015,Dvurechenskii,tianyi2020,janati2020}. For the purpose of our approach, we focus on free support barycenter with fixed weights in the following.

\section{\texttt{BaryScore} Metric} 
The construction of automatic metrics usually relies on two paradigms depending on the availability of a reference sentence for each candidate \cite{specia2010machine}. Here and throughout the paper, we assume that at least one reference is available for each candidate. Denote by $C=\{\omega_{\scriptscriptstyle 1}^{c},\ldots, \omega^{c}_{\scriptscriptstyle n_c} \}$ the candidate and $R=\{\omega_{\scriptscriptstyle 1}^{r},\ldots, \omega^{r}_{\scriptscriptstyle n_r} \}$ the reference composed of $n_c$ and $n_r$ words, respectively. Our goal is to design a metric $m: (C,R)\mapsto m(C,R) \in \mathbb{R}_+$ such that the closer to zero the better candidate is. 
\begin{figure*}
 \centering
\includegraphics[width=\textwidth]{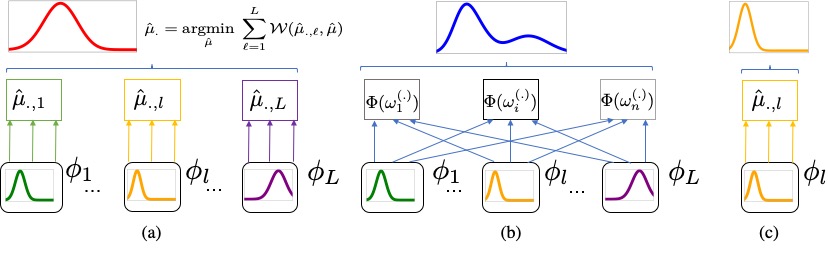}
 \caption{Schema of the different layers aggregation step: (a) Wasserstein barycenter (as used in \texttt{BaryScore}), (b) Power mean (as used in MoverScore), (c) best layer selection (case of BertScore). $\hat{\mu}_{\cdot,\ell}\quad \ell \in [1,L]$ stands for either the empirical distribution of the reference $R$ or the candidate $C$ text. Similarly, for words denoted by $\omega_i^{\cdot}$. }
\end{figure*}

\noindent\textbf{Algorithm.} Our metric $m$, named \texttt{Baryscore}, can  be summarized in two steps:

(i) Find the Wasserstein barycentric distributions of contextual encoder layers for $C$ and $R$;

(ii) Evaluate these barycentric distributions using the Wasserstein distance.

\noindent \textbf{Wasserstein barycenters.} Assume that a contextual encoder, (\textit{e.g.} BERT and ELMo), is composed of $L$ layers, \textit{i.e.},  $\phi_1,\ldots, \phi_L $ functions that map a candidate text C and a reference text R respectively  into $\phi_{\ell}({C}) \in \mathbb{R}^{ n_{c} \times d}$ and  $\phi_{\ell}({R}) \in \mathbb{R}^{  n_{r} \times d}$, for every $1\leq \ell \leq L$. In our approach, we consider the discrete probability distributions induced by $\phi_{\ell}({C})$ and $\phi_{\ell}({R})$, where $\phi_{\ell}({C})_{i}$ and $\phi_{\ell}({R})_{j}$ represent the embedding of the i-th token and j-th token of the candidate and reference text, respectively. Precisely, $2L$ empirical measures are constructed from these layers functions such that $\hat{\mu}_{C,\ell}= \sum_{i=1}^{n_c}\alpha_{i} \delta_{\scriptsize \phi_{\ell}(\omega_i^{c})}$ and $\hat{\mu}_{R,\ell}=  \sum_{j=1}^{n_r} \beta_{j}\delta_{ \scriptsize \phi_{\ell}(\omega_j^{r})}$, where $\alpha=\{\alpha_{1},\ldots, \alpha_{n_c} \}$ and $\beta=\{\beta_{1},\ldots, \beta_{n_r} \}$ are the vector of inverse document frequencies of each word $\omega_i$ of $C$ and $R$, respectively, and $\delta_x$ is the dirac mass at point $x$. Further, Wasserstein barycenters (see Eq.  \eqref{OT-barycenter}) are computed on the candidate C and the reference R leading to two barycentric embedding measures $\hat{\mu}_C$ and $\hat{\mu}_R$ with fixed sizes $n_c$ and $n_r$, respectively. Considering weights of the barycentric measures as uniform, as for the layers weights,  the optimization problem is equivalent to find locations  $\{x_i^c\}_{i=1}^{n_c}$ such that:  
\begin{equation}
\underset{ (x_1^{c},\dots, x_{n_c}^{c}) \in \mathbb{R}^d}{\text{argmin}} 
 \; \;   \sum_{\ell=1}^{L}\mathcal{W}(\hat{\mu}_{c,\ell}, \hat{\mu}_c)
\end{equation} 
with $\hat{\mu}_c = \frac{1}{n_c} \sum\limits_{i=1}^{n_c} \delta_{x_i^{c}}$. The previous formulation is similar for the reference text, replacing $C$ by $R$ and $\alpha$ by $\beta$ in notations. The final embedding, denoted by $\Phi$, is then considered as the locations of the barycentric measures, \textit{i.e., }
$\Phi(C) = \{x_1^{c}, \ldots, x_{n_c}^{c} \}$ and $\Phi(R) = \{x_1^{r}, \ldots, x_{n_r}^{r} \}$.

\noindent \textbf{Computing the Wasserstein distance.} The last step of our approach is then to evaluate discrete measures induced by the final embeddings, \textit{i.e., } the candidate and reference barycentric measures $\hat{\mu}_C$ and $\hat{\mu}_R$, using the Wasserstein distance, leading to the \texttt{Baryscore} given by $m(C,R)= \mathcal{W}(\hat{\mu}_C,\hat{\mu}_R)$. This step boils down to computing the WMD and is similar to the final step in MoverScore or SentenceMover. The entire procedure is summarized in Algorithm~\autoref{alg:optimization}.

\begin{algorithm}[h]
 \begin{algorithmic}
\INPUT $C=\{\omega_{\scriptscriptstyle 1}^{c},\ldots, \omega^{c}_{\scriptscriptstyle n_c} \}$, $R=\{\omega_{\scriptscriptstyle 1}^{r},\ldots,$ $ \omega^{r}_{\scriptscriptstyle n_r} \}$, $(\phi_1,\ldots,\phi_L)$ pre-trained layers from BERT or ELMo.

\noindent\textbf{Compute layers embeddings:} \\

$\phi_{\ell}(C)$ and $\phi_{\ell} (R)$ for every $1\leq \ell \leq L$.
\vspace{0.2cm}

\noindent\textbf{Compute measures:} $\{\hat{\mu}_{C,\ell}, \hat{\mu}_{R,\ell}\}_{\ell=1}^L$.  \\
\vspace{0.2cm}

\noindent\textbf{Compute Wasserstein barycenters:}
\begin{align*}
\hat{\mu}_C & = \underset{ \hat{\mu}}{\text{argmin}} \;  \sum_{\ell=1}^{L} \mathcal{W}(\hat{\mu}_{C,\ell}, \hat{\mu}),\\
\hat{\mu}_R & = \underset{ \hat{\mu}}{\text{argmin}} \;  \sum_{\ell=1}^{L} \mathcal{W}(\hat{\mu}_{R,\ell}. \hat{\mu}),
\end{align*}

\OUTPUT  $\mathcal{W}(\hat{\mu}_{R}, \hat{\mu}_C)$.
\end{algorithmic}
\caption{\texttt{BaryScore}}
\label{alg:optimization}
\end{algorithm}

\noindent\textbf{Parameters of BaryScore metric.} Our metric is dependant on the choice of the continuous representations (\textit{e.g.}, BERT, ELMo and Roberta) and then its performance will be  influenced by the choice of the model. As it is common in concurrent work \cite{zhang2019bertscore,zhao2019moverscore,clark-etal-2019-sentence} all the results in the paper are obtained with one single model: namely BERT-base-uncased. Additionally, we report results obtained with the BERT fine-tuned on NLI release in \cite{zhao2019moverscore}, this model is referred as \texttt{BaryScore}\textsuperscript{+} in the following. 
In contrast to previous work (\textit{e.g} SentenceMover) that integrates a preprocessing step by removing the stopwords based on a static list, we keep all words. Also, our framework provides a natural way to exploit all available layers of the model while previous work relies on a specific subset of them (\textit{e.g.},  MoverScore and BertScore).  We believe this strengthens the robustness of our approach.

\noindent \textbf{Comparison with the Moverscore.}
Following the footstep of \cite{zhang2019bertscore}, the Moverscore \cite{zhao2019moverscore}  applied optimal transport to the output of Contextualized Encoders (CE) such as BERT or ELMo. Precisely, let's assume that a CE is composed of $L$ layers, the Moverscore's context representation  is given for each word $\omega_j$ by
  $\Phi(\omega_j)=T(\phi_1(\omega_j),\ldots, \phi_L(\omega_j))$, where the transformation $T$ is either power means \cite{ruckle} or aggregation routines depicted in \cite{zhao2018,zhao2019}. The score is then defined by the Wasserstein distance between the empirical distributions given by $\Phi(C)$ and $\Phi(R)$.
  
 The main weakness of this approach is the aggregation step. Taking into account the role of the underlying geometry of the probability distribution  as well as the interpretability of the transportation flow are high benefits of Optimal transport.   However, performing Wasserstein distance after applying power means, \textit{i.e., }  an aggregation in an euclidian space (see \textit{e.g.},  \autoref{fig:barycenter}), does not allow a proper evaluation of the geometry induced by the CE layers in  the Wasserstein space. Indeed, Moverscore evaluates a distorted geometry inducing wrong interpretability of the transportation flow.  The advantage of exploiting  Wasserstein barycenter over euclidean aggregation relies on rehabilitating this geometry, as  shown in Section~\ref{sec:6}.

\section{Experimental Settings}
In this section, we present our evaluation methods as well as the various dataset used to benchmark our metric.

\noindent\textbf{Extension of notations.} In the previous section, we have only  considered a candidate and a reference sentence. In order to evaluate and compare different metrics, we need to extend the previous notations to include the system that generates each sentence. To this end, we will assume that we have a dataset: $\mathcal{D} = \{R_i, \{C_i^s,h(C_i^s)\}_{s=1}^S \}_{i=1}^N$, where $C^j_i$ is the $i$-th text generated by the $j$-th system, $h(C^j_i)$ is human score assign to $C^j_i$ and $R_i$ the reference text associated to $C^j_i$;  $N$ is the number of available texts; S the number of different systems. 
\subsection{Evaluating automatic evaluation of NLG} 
The quality of the evaluation metric is measured by its correlation with the human judgment \cite{chatzikoumi2020evaluate,specia2010machine,koehn2009statistical,banerjee2005meteor}. Three correlation measures can be considered: Pearson \cite{leusch2003novel}, Spearman \cite{melamed2003precision} or Kendall \cite{kendall1938new}. In addition, two different levels of granularity are considered to compute theses correlation coefficients.

\noindent \textbf{System level correlation.} These can be considered when assessing the discrimination capability between two systems. This level of correlation tries to answer the question: ``\textit{Can the metric be used to compare the performance of two systems?}''. Formally, the system level correlation $K_{sys}$ measures the quality of a metric $m$ defined as:  
\begin{align}\label{sy}
    K_{sys} =& K(M^{sy}, H^{sy}),\\
    M^{sy} =&\hspace{-0.1em} \bigg{[}\hspace{-0.1em}\frac{1}{N}\hspace{-0.1em}\sum_{i=1}^N\hspace{-0.1em} m(R_i,C^1_i)\hspace{-0.1em},\hspace{-0.1em}\cdots\hspace{-0.1em},\hspace{-0.1em} \frac{1}{N}\hspace{-0.1em}\sum_{i=1}^n\hspace{-0.1em}m(R_i,C^s_i)\hspace{-0.1em}\bigg{]}\hspace{-0.1em},\nonumber
    \\
    H^{sy} =&\hspace{-0.1em} \bigg{[}\frac{1}{N}\sum_{i=1}^N {h}(C^1_i),\cdots,\frac{1}{N}\sum_{i=1}^N {h}(C^S_i)\bigg{]},\nonumber
\end{align}
where K is the considered correlation coefficient.

\noindent\textbf{Text level correlation.} This is computed to evaluate the ability of a metric to measure the semantic equivalence between a candidate and a reference sentence. Such a level of correlation aims at providing an answer to the question: ``\textit{Can the metric be used as a loss or reward of a system?}''. By introducing similar notations to  those in \autoref{sy}, we obtain the text level correlation $K_{text}$:
\begin{align} \label{se}
    K_{text} =&  \frac{1}{N} \sum_{i=1}^N K ( M^{text}_i, H^{text}_i ), \\
    M^{text}_i =& \big{[}m(R_i,C^1_i),\cdots,m(R_i,C^S_i)\big{]} ,\nonumber \\
    H^{text}_i =& \big{[}{h}(C^1_i),\cdots,{h}(C^S_i)\big{]}.\nonumber
\end{align}

\noindent\textbf{Significance testing.} To ensure that observed improvement is statistically significant we follow common consensus \cite{deutsch2021statistical,graham2015improving,graham2015accurate,graham2014testing} in NMT and rely on William test \cite{steiger1980tests} as considered observations are correlated\footnote{An example code is provided by the authors at \url{https://github.com/ygraham/nlp-williams}}. 

\subsection{Choice of datasets}
We motivate our choice of datasets for each tasks.

\noindent\textbf{Translation.} Multiple translation datasets are available from the WMT translation shared tasks \cite{bojar-etal-2014-findings,bojar-etal-2015-findings,bojar-etal-2016-findings,bojar-etal-2017-findings}. 
Keeping in mind the work by \cite{card2020little} stressing the importance of the size of the dataset, we follow \cite{zhang2019bertscore} and choose to work with WMT16 and additionally report results on WMT15 that both offer over 500 sentences per language (in contrast to new versions \cite{,bojar-etal-2018-findings,bojar-etal-2019-findings,bojar-etal-2020-findings} that rely on a lower number--around $50$--of annotated texts).

\noindent\textbf{Summarization.} Several datasets have been introduced to compare metrics for summarization. Classical choices include MSR Abstractive Text Compression dataset \cite{toutanova2016dataset}, TAC datasets \cite{dang2008overview,mcnamee2009overview} or on news summarization from CNN/DailyMail \cite{hermann2015teaching,nallapati2016abstractive}. TAC datasets and contains flaws \cite{rankel2013decade,peyrard2019studying,bhandari2020re}, thus in this work we rely on the CNN introduced in \cite{bhandari2020re}. It is composed of 11,490 summaries comming from 11 extractive systems \cite{lewis2019bart,yoon2020learning,raffel2019exploring,gehrmann-etal-2018-bottom,dong2019unified,liu2019text,chen2018fast,see2017get} and 14 abstractive systems \cite{zhong2020extractive,wang2020heterogeneous,zhong2019searching,liu2019text,zhou2018neural,narayan2018ranking,dong2019unified,kedzie2018content,zhou2018neural}.

\noindent\textbf{Data2Text.} In contrast to previous work that rely on old task-oriented dialogue datasets (\textit{i.e.},  BAGEL \cite{mairesse2010phrase}, SFHOTEL \cite{wen2015semantically}), we focus on the WebNLG challenge \cite{perez2016building,gardent2017creating,ferreira2018enriching,ferreira20202020} as sentence available in this challenge is more representative of progress of NLG\footnote{System description and performance are available at \url{https://webnlg-challenge.loria.fr/}}. The data provides from 15 systems relying on either symbolic or neural approaches.  In this challenge, multiple evaluation criteria are used to assess the quality of the text. Due to the space limitations, we compute correlation on the three following criteria: (1) \textit{data coverage} which measures if the generated text contains all the available information present in the input data, (2) \textit{relevance} which characterizes if the generated text is solely composed with information available in the input, (3) \textit{correctness} which measures if all input information is both correct and adequately introduced.

\noindent\textbf{Image captioning.} To evaluate metrics on image captioning, there is a consensus to exploit COCO datasets \cite{lin2014microsoft} and to compute correlation at the system level. On average each image has five reference captions. Each of the 12 systems is evaluated on 5 criteria. Following \cite{anderson2016spice,zhao2019moverscore,zhang2019bertscore}, we only compute correlation on M1, M2 which are related to the overall quality of the caption. 


\section{Numerical Results} \label{sec:6}
In this section, we study the performance of \texttt{BaryScore} on the four aforementioned tasks.
\begin{figure}
 \centering
    \includegraphics[width=0.25\textwidth]{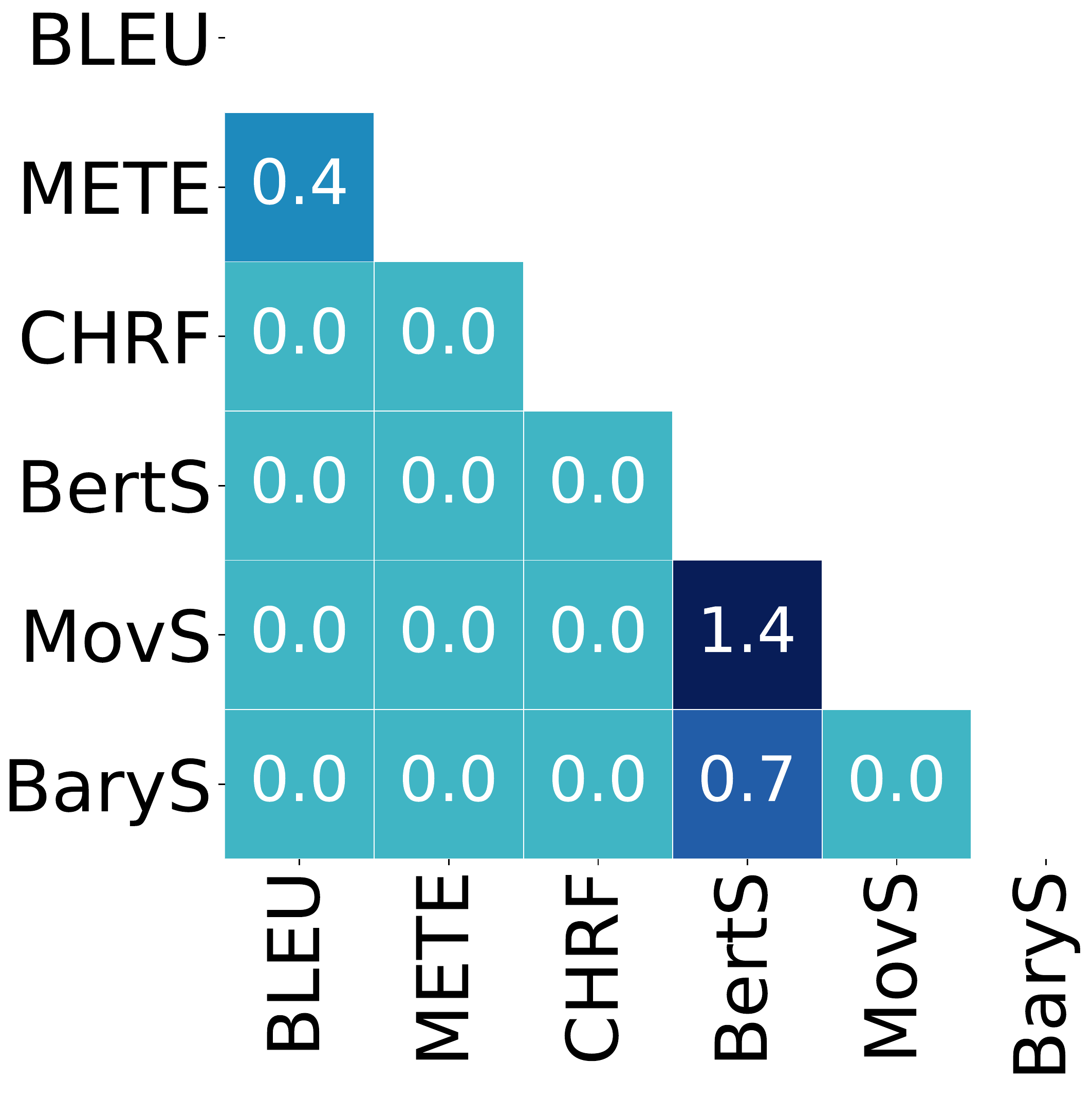}
 \caption{Significance testing on de-en for WMT16. In the matrix is reported the p-value of the William Test in percent. }
    \label{fig:significance}
\end{figure}
\subsection{Translation}

\begin{table*}
 \centering
 \resizebox{0.8\textwidth}{!}{\begin{tabular}{lcccccccccccc}  \toprule {} 
 & \multicolumn{3}{c}{cs - en} & \multicolumn{3}{c}{de - en} & \multicolumn{3}{c}{ru - en} &\multicolumn{3}{c}{fi - en} \\
      \cmidrule(lr){2-4} \cmidrule(lr){5-7}  \cmidrule(lr){8-10} \cmidrule(lr){11-13}  
 &  $r$  &  $\rho$ &  $\tau$ &  $r$ &  $\rho$  & $\tau$ & $r$ &  $\rho$ & $\tau$ & $r$ &  $\rho$ $\tau$ & \\
      \cmidrule(lr){2-4}\cmidrule(lr){5-7}  \cmidrule(lr){8-10} \cmidrule(lr){11-13}   
 \texttt{BaryS}\textsuperscript{+}  &       \textbf{75.9} &        \textbf{75.2} &  \textbf{56.9}        &       \textbf{75.8} &        \textbf{74.5}  & \textbf{56.2} &       \textbf{77.6} &        \textbf{75.0} & \textbf{56.7} &       \textbf{79.9} &        \textbf{78.7} & \textbf{59.7}  \\ 
  \texttt{BaryS} &       74.2 &        73.8 &   54.9      &       74.1 &        72.2 & 53.1 &       73.7 &        70.6  & 53.2&       76.6 &        74.5 & 56.3  \\ 
 \texttt{BertS}-F                                     &       74.3 &        73.5 &    54.3     &       72.2 &        70.7 & 52.9 &       74.0 &        70.5 & 52.5 &       74.7 &        72.5 & 54.1  \\ 

  \texttt{MoverS}\textsuperscript{+}                                       &       71.0 &        71.1 &  52.0     &         71.1 &        68.2 & 50.3 &       67.3 &        64.7 & 47.3 &       72.2 &        72.0  & 52.9 \\ 
 \texttt{MoverS}                              &       68.8 &        70.1 &  52.0    &          71.8 &        69.4 & 50.3 &       68.6 &        65.5 &   46.9&       70.0 &        70.0  & 52.3 \\ 
\texttt{MET}                                         &       56.4 &        57.1 &  40.9   &           60.1 &        61.4 & 61.5 &       58.9 &        59.6 & 42.4 &       58.8 &        59.0  & 42.9 \\ 
 \texttt{BLEU}                                                &       44.5 &        44.5 & 31.1     &           53.6 &        48.0  & 48.1&       53.4 &        49.0  & 34.1&       46.5 &        41.8  & 29.3 \\ 
 \texttt{CHRF}                                               &       26.0 &        21.6 &  15.8    &          29.5 &        28.9 & 28.9 &       32.5 &        32.9 & 23.4 &       30.2 &        26.5 & 19.2  \\ \hline  
 \end{tabular}}
 \caption{Absolute correlations between metric prediction and text level human judgement on 4 pairs of WMT15.}\label{tab:wm15}
 \end{table*}
\begin{table*}
 \centering
 \resizebox{\textwidth}{!}{\begin{tabular}{lcccccccccccccccccc}  \hline {} 
  & \multicolumn{3}{c}{cs - en} & \multicolumn{3}{c}{de - en} & \multicolumn{3}{c}{ru - en} &\multicolumn{3}{c}{fi - en}
  &\multicolumn{3}{c}{ro - en}
  &\multicolumn{3}{c}{tr - en}\\
      \cmidrule(lr){2-4} \cmidrule(lr){5-7}  \cmidrule(lr){8-10} \cmidrule(lr){11-13} 
      \cmidrule(lr){14-16} \cmidrule(lr){17-19} 
 & $r$ &  $\rho$ & $\tau$ &  $r$ &  $\rho$ & $\tau$ &  $r$ &  $\rho$  & $\tau$&  $r$ &  $\rho$ & $\tau$ &  $r$ &  $\rho$ & $\tau$ &  $r$ &  $\rho$ & $\tau$ \\
      \cmidrule(lr){2-4} \cmidrule(lr){5-7}  \cmidrule(lr){8-10} \cmidrule(lr){11-13} 
      \cmidrule(lr){14-16} \cmidrule(lr){17-19} 
 \texttt{BaryS}\textsuperscript{+}  &  
 \textbf{76.6} &        \textbf{76.2}& \textbf{57.5}
 &       \textbf{68.5} &        \textbf{67.7}& \textbf{50.0}
 &       \textbf{69.4} &        \textbf{68.3}& \textbf{51.3} 
 &       \textbf{70.2} &        \textbf{69.5}& \textbf{50.9} 
 &       \textbf{74.3} &        \textbf{73.0}& \textbf{54.5} 
 &       \textbf{73.8} &        \textbf{70.5}& \textbf{52.4}  \\ 
 \texttt{BaryS}  
 &       74.2 &        74.3& 56.3 &
 64.6 &        64.2& 47.9 &
 67.5 &        66.4 & 48.1& 
 67.1 &        66.4& 48.3 & 
 72.5 &        71.4& 52.9 &     
 69.3 &        67.1   & 51.4\\ 
\texttt{BertS}-F                                    
&       74.1 &        73.6& 55.49
&       65.3 &        64.6& 46.3 &   
65.1 &        64.6& 46.9 &  
65.4 &        64.1& 47.0 &   
70.2 &        67.6& 49.5 &    
70.7 &        67.1  & 49.0  \\  
\texttt{MoverS}\textsuperscript{+}         
&       70.7 &        70.4& 53.4 &  
62.4 &        60.7& 44.8 &     
64.0 &        62.2 & 45.2&     
64.5 &        62.6& 45.8 &   
66.4 &        66.0& 48.6 &  
66.3 &        60.7  & 44.9  \\ 
\texttt{MoverS}                         
&       67.4 &        69.5& 52.6 
&       60.9 &        59.1& 44.2
&       64.4 &        62.8& 44.8
&       63.1 &        62.2& 45.1
&       64.2 &        65.4& 48.2
&       66.1 &        64.0& 43.7   \\ 
\texttt{MET}                                       
&       64.5 &        67.2& 49.2 
&       51.6 &        50.0& 35.3 
&       54.8 &        57.5& 41.2 
&       53.9 &        52.4& 37.3
&       58.6 &        59.4& 42.1
&       61.8 &        59.1& 42.4  \\ 
\texttt{BLEU}                                            
&       53.8 &        52.3& 36.2
&       45.3 &        40.8& 28.2
&       46.3 &        43.8& 30.2 
&       39.9 &        37.7& 26.3 
&       47.0 &        43.2& 30.2
&       47.1 &        43.9 & 30.4  \\ 
\texttt{CHRF}                                              
&       24.4 &        26.8& 19.2 
&       34.2 &        33.8& 24.3 
&       28.7 &        29.8& 21.2
&       14.4 &        15.3& 11.2 
&       16.0 &        12.6& 9.0
&       26.5 &        16.2 & 12.4  \\ \hline  \end{tabular}}
 \caption{Absolute correlations between metric prediction and text level human judgement on 6 pairs of WMT16.}\label{tab:wm16}
\end{table*}

\begin{figure*}[!htb]
\centering
\begin{minipage}{.29\textwidth}
\vspace{2.2em}
    \includegraphics[width=\textwidth]{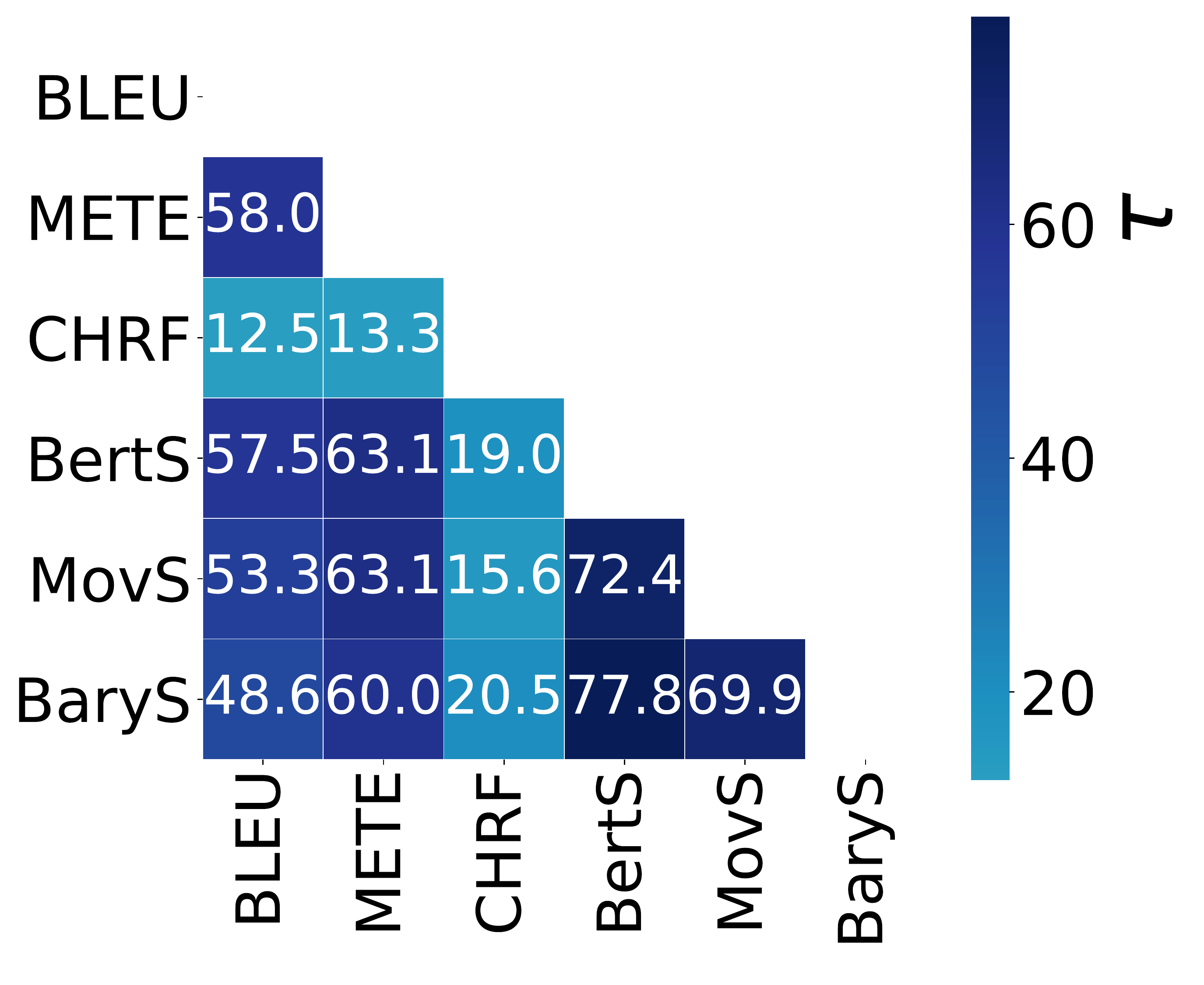}
 \caption{de-en WMT16 Kendall correlation between predictions made by different metrics. }\label{fig:correlation_matrix}
\end{minipage}\hspace{0.01\textwidth}\begin{minipage}{.7\textwidth}
    \resizebox{\textwidth}{!}{
\begin{tabular}{ccccccc||cccccc}\hline
    & \multicolumn{6}{c}{Abstractive} & \multicolumn{6}{c}{Extractive} \\
 \cmidrule(lr){2-7} \cmidrule(lr){8-13}
    & \multicolumn{3}{c}{Text}  & \multicolumn{3}{c}{System}  
    & \multicolumn{3}{c}{Text}  & \multicolumn{3}{c}{System} \\
     \cmidrule(lr){2-4} \cmidrule(lr){5-7}  \cmidrule(lr){8-10} \cmidrule(lr){11-13}
 & $r$ & $\tau$	 & $\rho$& $r$ & $\tau$	 & $\rho$ & $r$ & $\tau$	 & $\rho$& $r$ & $\tau$	 & $\rho$ \\
     \cmidrule(lr){2-4} \cmidrule(lr){5-7}  \cmidrule(lr){8-10} \cmidrule(lr){11-13}

\texttt{BaryS}	  &	\textbf{75.6} &	\textbf{75.9} &	\textbf{76.6}& 94.9	&  83.4& 	\textbf{96.3}  & 81.1&	81.1&	70.7  &  63.2 &	41.8 &	45.4 \\
\texttt{BaryS}\textsuperscript{+}   &	72.4 &	73.4 &	76.3& 95.9& 	83.4&	96.2  & 	\textbf{81.3}&	\textbf{81.5}&	\textbf{70.9} &  64.0 &	38.1 &	48.1  \\
 \texttt{BertS}-F &	73.4 &	74.9 &	70.6& 92.3 &	82.4 &	93.4  & 	29.4	&18.9&	55.0 &  69.0 &	49.0 &	62.7  \\
 \texttt{BertS}-R  &	71.7 &	71.9 &	72.0& 92.3 &	82.4 &	93.4  & 	70.9&	72.9&	73.8 &  06.9&	01.8& 	02.7   \\
\texttt{MoverS}  &	72.4 &	71.9 &	73.0& 82.8 &	75.8 &	92.5  & 76.1 &	76.1 &	47.4 &  16.1& 	09.0& 	09.0  \\
\texttt{MoverS}\textsuperscript{+}  &	73.3 &	71.2 &82.8 &	75.8 &	92.5  &	73.0& 	76.4 &	76.0 &	47.0 & 21.1& 	15.1& 	29.0 \\
Rouge-1 &	73.5 &	73.0 &	74.5 &90.3 &	80.2 &	92.5  & 	72.2&	74.0&	69.1 &  \textbf{73.2} &	\textbf{52.7} &	\textbf{69.0}  \\
Rouge-2	&	73.0 &	73.5 &	73.0 &\textbf{98.5} &	\textbf{89.0} &	96.0  & 	55.1&	53.2&	69.0 &  69.9 &	41.8 &	54.5  \\
\texttt{JS-2} & 68.9 &.6.8 &	69.8 & 92.5 &	82.4 &	92.9 & 05.5	& 12.7& 	19.0 &  5.5& 	12.7	& 19.0  \\
\hline
    \end{tabular}}
    \caption{Absolute correlation coefficients (as measured using Pearson ($r$), Spearman ($\rho$) and Kendall ($\tau$) coefficient) between different metrics on text summarization.}
    \label{tab:summarization_results}
\end{minipage}
\end{figure*}

\textbf{Overall results.}
\autoref{tab:wm15} and \autoref{tab:wm16} gather correlation to human judgments on WMT15 and WMT16. We conduct a statistical analysis to ensure that the observed improvements are statistically meaningful (see \autoref{fig:significance}).

We observe that \texttt{BaryScore\textsuperscript{+}} is the best performing metric on both datasets for all languages. Similarly to \cite{zhao2019moverscore}, we observe an improvement when using their pretrained version of BERT on MNLI \cite{glue}. By comparing the best performance achieved by \texttt{BaryScore} compared to \texttt{MoverScore}, we hypothesize  that Wasserstein barycenter preserves more geometric properties of the information learnt by BERT. 
\\\textbf{Correlation analysis.} \autoref{fig:correlation_matrix} reports the inter-correlation across metrics according to the Kendall $\tau$. We observe that the metrics based on BERT (\textit{e.g} BertScore, MoverScore and \texttt{BaryScore}) obtain medium-high correlation demonstrating that both the aggregation mechanism (\textit{e.g.},  one layer selection, power mean or Wasserstein barycenter) as well as the choice of similarity metric (\textit{e.g.}, cosine similarity, Wasserstein distance) affects the ranking of the predictions. 
\\\textbf{Takeaways:} Overall \texttt{BaryScore} is particularly suitable to compare two examples and thus could be used as an alternative to the standard cross-entropy loss to train NMT systems. On the other hand, our implementation made based on POT \cite{flamary2017pot} makes the speed comparable with MoverScore. We are able to process over $180$ sentence pairs  per second with \texttt{BaryScore} compared to $195$ sentence pairs per second with MoverScore on an NVIDIA-V100 GPU.
\subsection{Summarization}
\autoref{tab:summarization_results} reports results on the summarization task. We are able to reproduce the performance reported in the original paper \cite{bhandari2020re}. Contrarily to MT, we observe that there is no metric that can outperform all others on all correlation measurements. We can also notice that the improvement induced by the BERT fine-tuned on MNLI is not observed on this dataset for both \texttt{BaryScore} and MoverScore. 
\begin{figure*}[!htb]
\centering
\begin{minipage}{.29\textwidth}

 \resizebox{0.88\textwidth}{!}{\begin{tabular}{lrr} \hline
  & M1 & M2 \\
      \cmidrule(lr){2-2} \cmidrule(lr){3-3}
     \texttt{BaryS}\textsuperscript{+} & 85.6 & 83.9 \\
      \texttt{BaryS}  & 85.2 & 82.6 \\
   \texttt{MoveS}\textsuperscript{+} & 83.1 & 82.6 \\
    \texttt{MoveS}  & 78.1 & 82.1 \\
  \texttt{BertS} - P & 83.0 & 82.1 \\
   \texttt{BertS} - R & 80.1 & 75.2 \\
    \texttt{BertS} - F & 79.1 & 81.1 \\\hline
         \texttt{BLEU}  & 58.3 & 60.8 \\
        \texttt{METEOR} & 60.3 & 59.9 \\ 
        \texttt{SPICE\textsuperscript{*}} & 75.9 & 75.0 \\\hline
        \texttt{CIDER\textsuperscript{*}} & 43.8 & 44.0 \\\hline
        
        \texttt{LEIC\textsuperscript{*}} & \textbf{93.9} & \textbf{94.9}\\\hline
\end{tabular}}\caption{System level Pearson correlation with human judgement on the MSCOCO dataset.}\label{tab:captionning}
\end{minipage}\hspace{0.01\textwidth}\begin{minipage}{.7\textwidth}
 \resizebox{\textwidth}{!}{\begin{tabular}{lrrrrrrrrr}\hline  
   &  \multicolumn{3}{c}{Correctness} & \multicolumn{3}{c}{Data Coverage} & \multicolumn{3}{c}{Relevance} \\
  Metric &  $r$ &  $\rho$ & $\tau$  &  $r$ &  $\rho$ &  $\tau$  &  $r$ &  $\rho$ &  $\tau$    \\
\midrule               Correct &   100.0    &    100.0    &    100.0   &    97.6 &     85.2 &    73.3 &        99.1 &     89.7 &    75.0     \\                 
DataC &      85.2 &       97.6 &      73.3 &  100.0   &    100.0  &   100.0  &        96.0 &     93.8 &    81.6 \\        
Relev &      89.7 &       99.1 &      75.0 &    96.0 &     93.8 &    81.6 &     100.0   &  100.0    &  100.0  \\       \hline

\texttt{BaryS} &\textbf{91.7} &\textbf{90.0}& 	\textbf{78.3}&			\textbf{87.8} &	78.2 &	61.6	& \textbf{89.4} &	\textbf{82.6} &	70.0 \\ 
\texttt{BaryS\textsuperscript{+}} & 90.5 &	89.5&	76.6 & 87.7	& 	\textbf{85.0}&  \textbf{70.0}	& 89.2 &	86.4 &	{71.6} 		\\
BertS &      {85.5} &       83.4 &      {73.3} &    
74.7 &     {68.2} &    53.3 &      
{83.3} &     {79.4} &    {65.0}  \\       
MoverS &      84.1 &       {84.1} &      {73.3} &    
{78.7} &     66.2 &    {53.3} & 
82.1 &     77.4 &    65.0  \\\hline  
BLEU &      77.6 &       66.3 &      60.0 &    
55.7 &     50.2 &    36.6 &    
63.0 &     65.2 &    51.6  \\
R-1 &      80.6 &       65.0 &      65.0 &  
76.5 &     76.3 &    60.3 &    
64.3 &     {69.2} &    56.7  \\
R-2 &      73.6 &       63.3 &      58.3 &    54.7 &     43.1 &    35.0 &    
62.0 &     60.8 &    46.7  \\
R-WE &      60.9 &       73.4 &      60.0 &    40.2 &     58.2 &    40.1 &    
49.9 &     64.1 &    48.3  \\
METEOR &      {86.5} &       {66.3} &      {70.0} &    
{77.3} &     50.2 &    46.6 &    
{82.1} &     65.2 &    58.6  \\
TER &      79.6 &       78.3 &      58.0 &    69.7 &     58.2 &    38.0 &   75.0 &     70.2 &    {\textbf{77.6}}  \\
\bottomrule\end{tabular}}
\caption{Correlation at the system level with human judgement along five different axis: correctness, data coverage, fluency, relevance and text structure for the WebNLG task. Overall best result is bolted.}
    \label{tab:web_nlg_sys}
\end{minipage}
\end{figure*}

\noindent \textbf{Consistency and robustness of \texttt{BaryScore}.} In contrast to what is observed on abstractive systems, we observe a strong inconsistency in the behavior of the previous metrics based on BERT for extractive systems. Indeed, at the text level BertS-R, MoverScore and MoverScore\textsuperscript{+} achieve good medium/high correlation whereas at the system level the achieved correlation collapses (correlation scores below 20 points). BertS-F, on the contrary, under-performs at the text level for extractive systems but achieves competitive performance with Rouge (the best performing metric) at the system level. We observe that using Wasserstein barycenter is a better way to aggregate the layer and provides better robustness as it alleviates the aforementioned problem. Indeed, the performance achieved by \texttt{BaryScore} is competitive at both the text and system levels.
\\\textbf{Takeaways:} Overall, the two versions of \texttt{BaryScore} are among the best performing metrics, outperform current BERT based metrics on 3/4 configurations, and achieve consistent performance on the 4th configuration. The consistent behavior of \texttt{BaryScore} demonstrates the validity of our approach for summarization. Whereas, a simpler and lighter alternative to \texttt{BaryScore}, as well as other BERT-based metrics to compare systems on summarization, remains the ROUGE score for 3/4 configurations. 

\subsection{Data2Text}
\autoref{tab:web_nlg_sys} reports results on data2text task using the WebNLG2020 data. To the best of our knowledge this is one of the first study using this dataset. \autoref{tab:web_nlg_sys} shows a strong correlation between the three evaluation dimensions with correlation $r$ and $\rho$ higher than 90. We observe that \texttt{BaryScore} consistently  metrics based on BERT and achieves best results on 6/9 configurations for \texttt{BaryScore} and 2/9 for \texttt{BaryScore}\textsuperscript{+}.

\subsection{Image Captioning}
We follow \cite{zhang2019bertscore,zhao2019moverscore} and report in \autoref{tab:captionning}\footnote{Results with * are reported from \cite{zhao2019moverscore}} Pearson correlation coefficients between prediction and system level judgment. Although  we were unable to reproduce exactly the results by  \cite{zhao2019moverscore}, we obtain comparable numbers and similar orderings. 
\\\textbf{Takeaways:} \texttt{BaryScore} outperforms current metrics except for \texttt{LEIC} that rely on information extracted from both image and text. These results validate the use of \texttt{BaryScore} to compare the performance of image captioning systems. 

\section{Summary and Concluding Remarks}
In this paper, we present a metric named \texttt{BaryScore} which relies on optimal transport and solves the geometric discrepancies present in existing metrics that use contextualized embedding with WMD. The present work is carried out in the context of NLG but it introduces a generic theoretically-grounded framework that could be extended to other NLP studies. In particular, it illustrates applications of Wasserstein barycenters to combine the different views offered by different layers of a deep neural network. Specifically, futur work includes testing Wasserstein Barycenters in a multimodal setting \cite{garcia-etal-2019-token,colombo2021improving}, for classification (\textit{e.g.} emotion \cite{DBLP:conf/wassa/WitonCMK18}, dialog act \cite{chapuis2020hierarchical,colombo2020guiding,DBLP:journals/corr/abs-2108-12465}, stance \cite{dinkar2020importance})  and controlling style in NLG \cite{jalalzai2020heavy,colombo-etal-2019-affect,DBLP:journals/corr/abs-2105-02685} .

\section{Acknowledgment}
Pierre is funded by IBM. This work was also granted access to the HPC resources of IDRIS under the project 2021-101838 made by GENCI.

\bibliographystyle{acl_natbib}
\bibliography{acl2021}

\clearpage
\section{Appendix}
\begin{figure*}[!htb]
\centering
    \begin{minipage}{.3\textwidth}
           \centering
    \subfloat{\includegraphics[width=\textwidth]{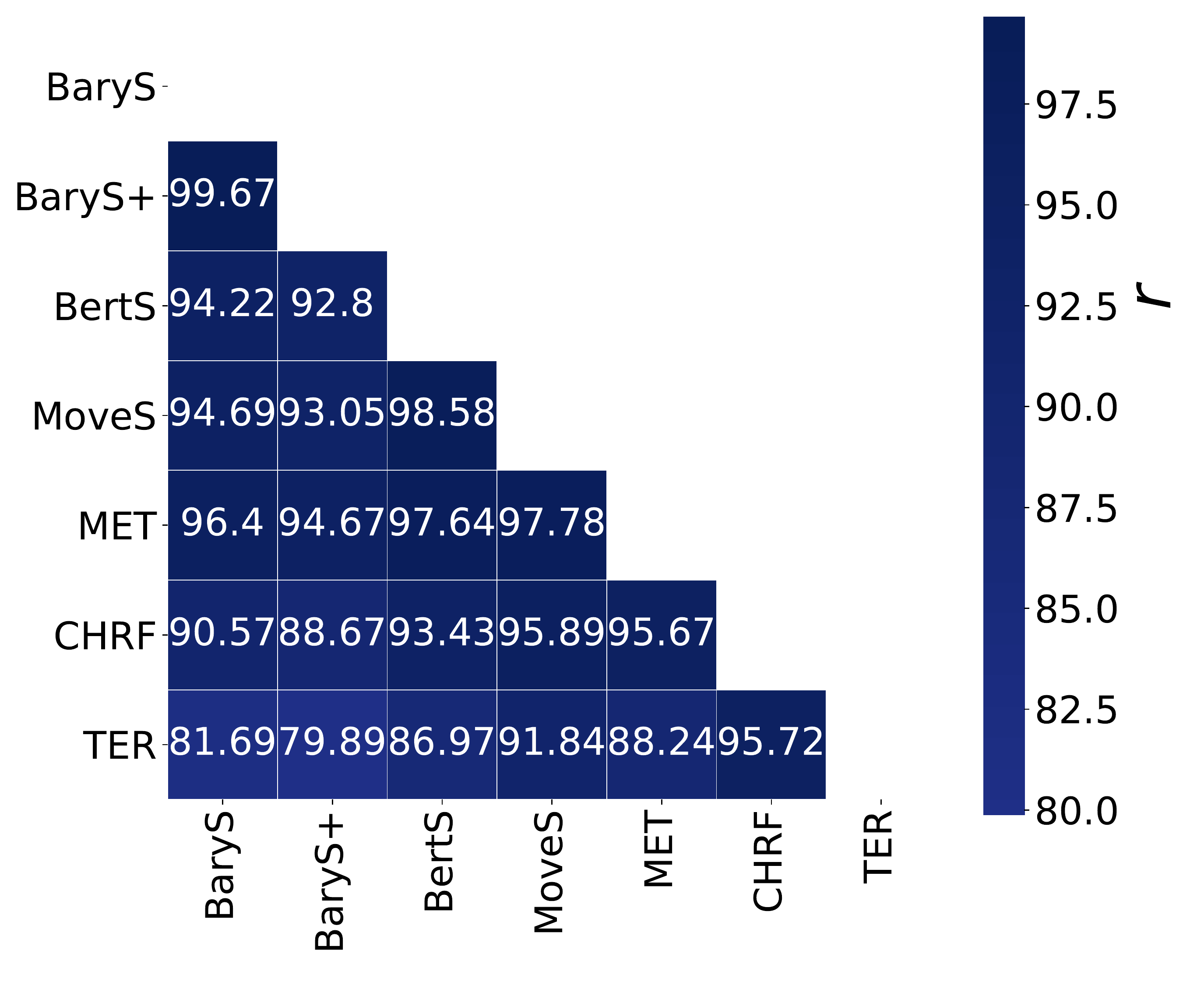} }
    \caption{Pearson $r$} 
    \end{minipage}
    \quad
    \begin{minipage}{.3\textwidth}
           \centering
    \subfloat{\includegraphics[width=\textwidth]{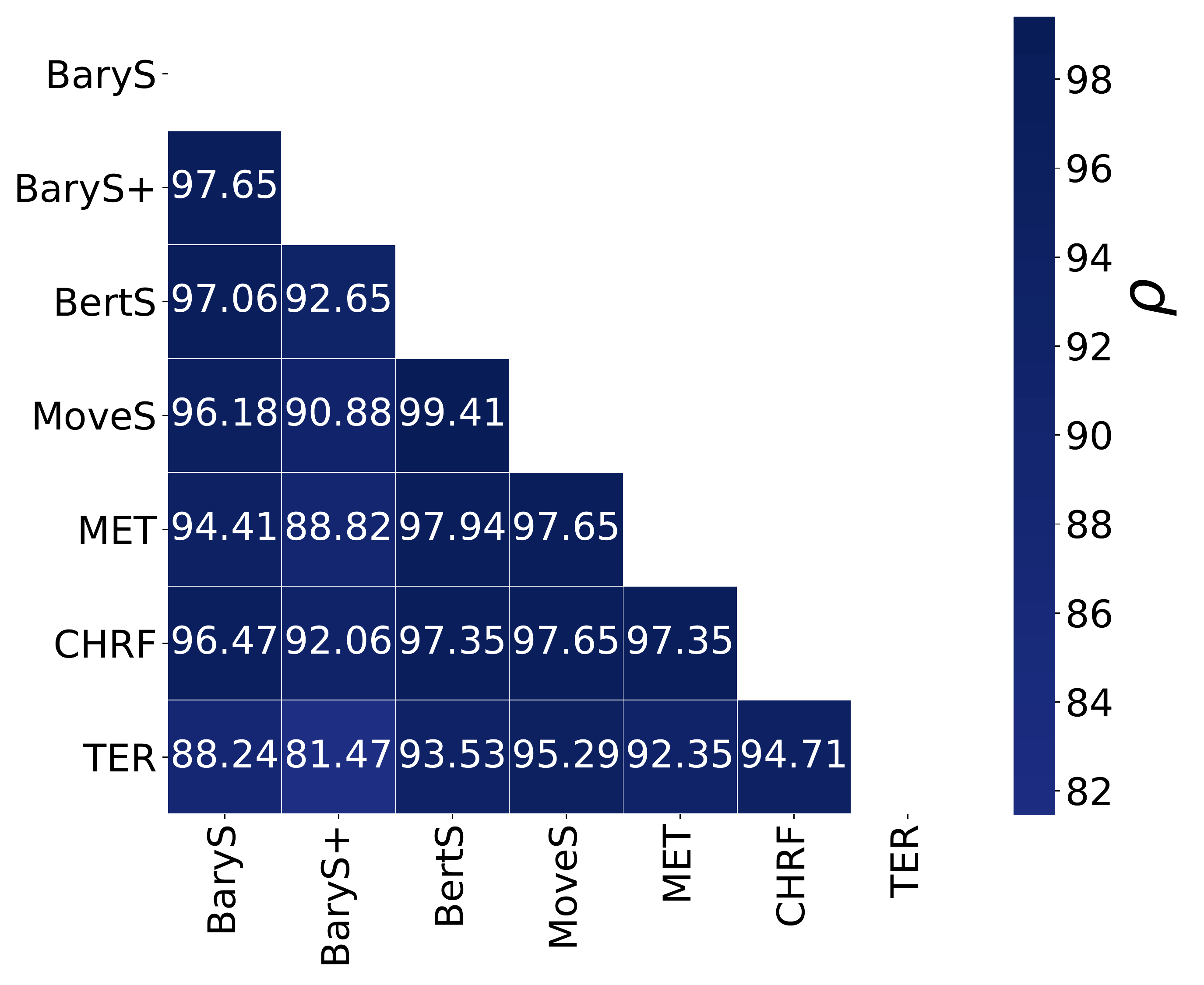} }
    \caption{Spearman $\rho$} 
    \end{minipage}
    \quad
    \begin{minipage}{.3\textwidth}
           \centering
    \subfloat{\includegraphics[width=\textwidth]{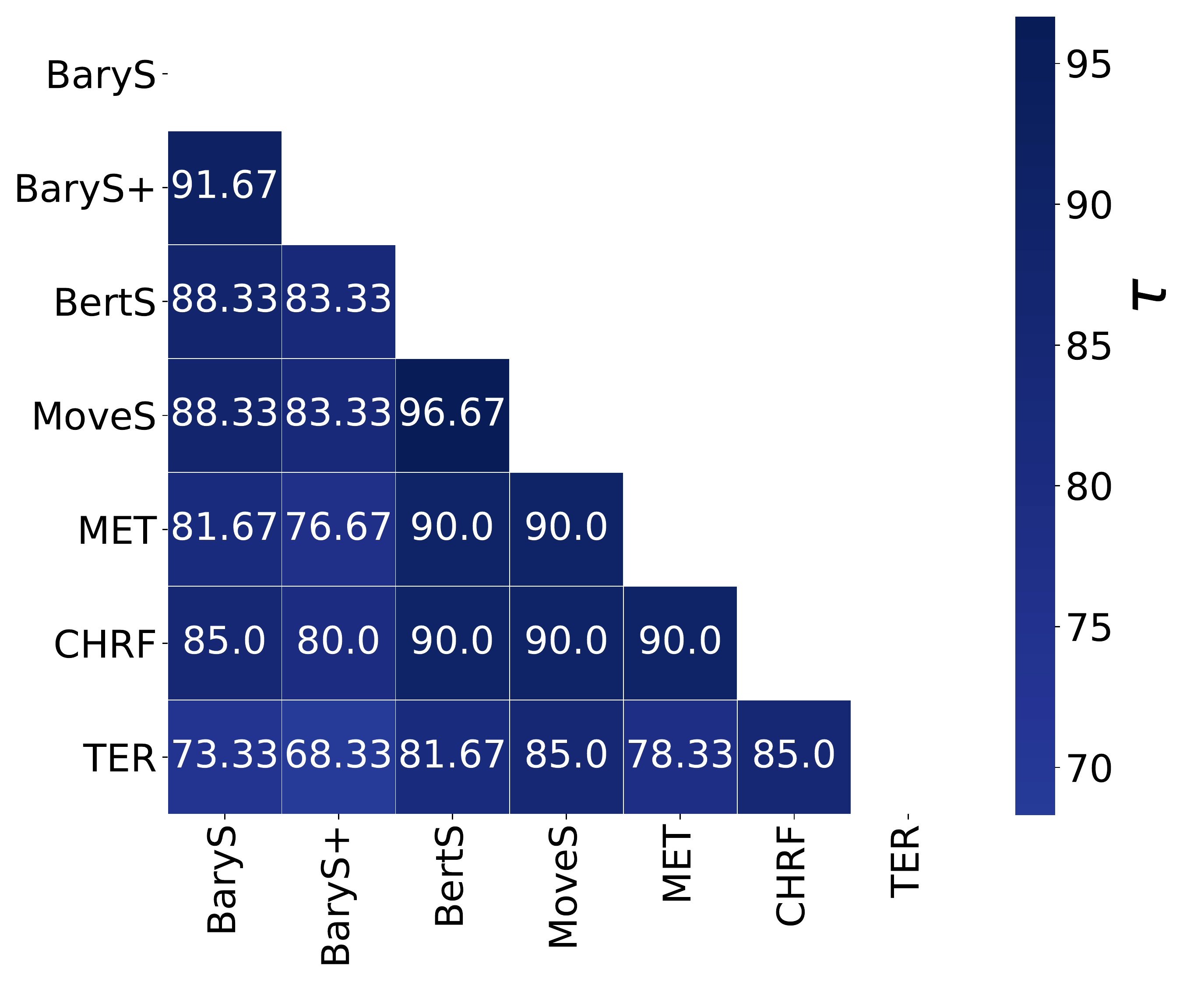} }
    \caption{Kendall $\tau$}
    \end{minipage}
    \caption{Correlation coefficients measuring correlation between the prediction made by different metrics based on BERT on WMT16 (de-en). }\label{fig:kendall_matrix_de_en}
\end{figure*}
\begin{figure*}[!htb]
\centering
    \begin{minipage}{.33\textwidth}
           \centering
    \subfloat{\includegraphics[width=\textwidth]{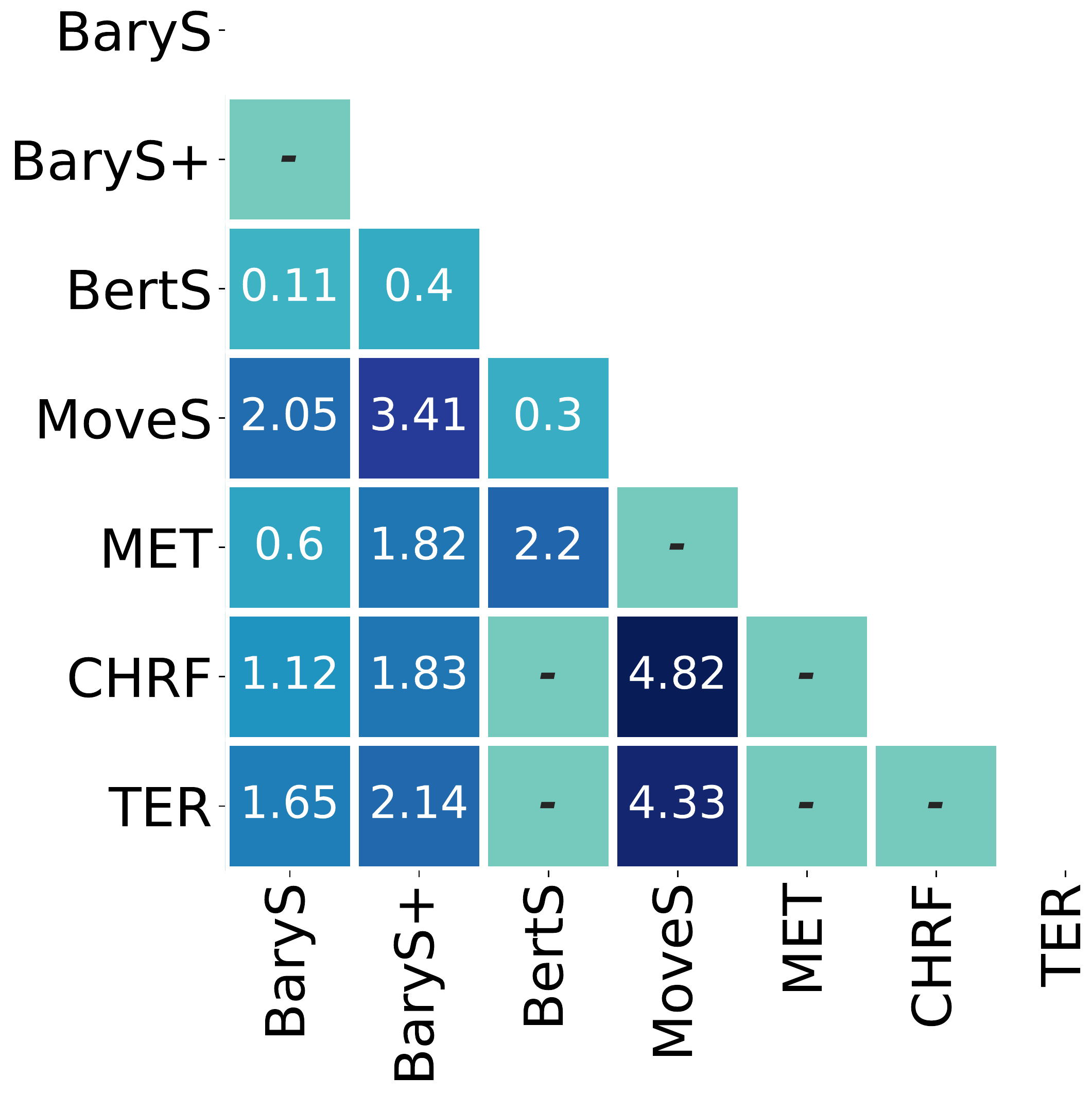} }
    \caption{Correctness}
    \end{minipage}
    \begin{minipage}{.3\textwidth}
    \subfloat{\includegraphics[width=0.9\textwidth]{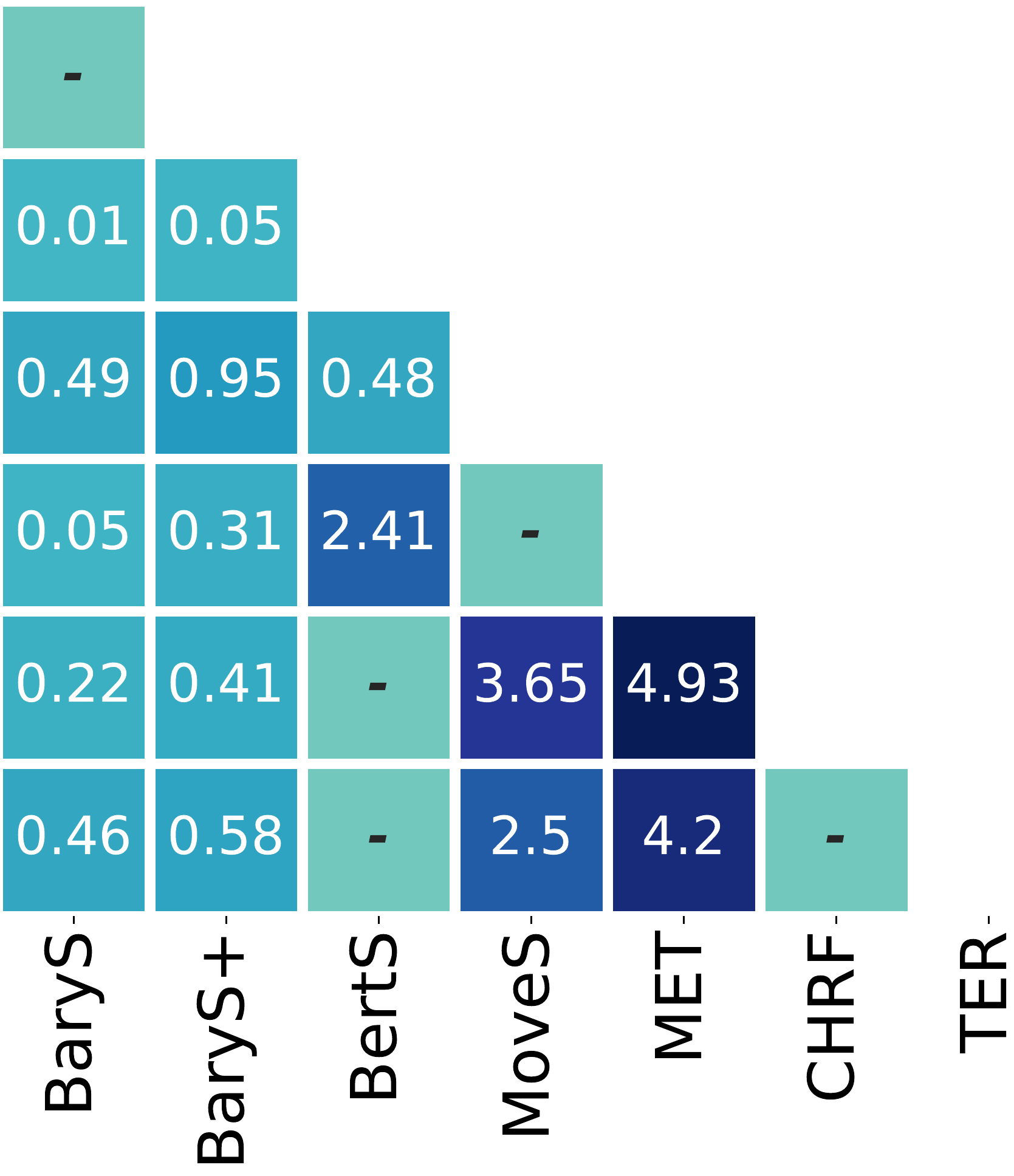} }
    \caption{Data Coverage}
    \end{minipage}
    \begin{minipage}{.3\textwidth}
    \subfloat{\includegraphics[width=0.9\textwidth]{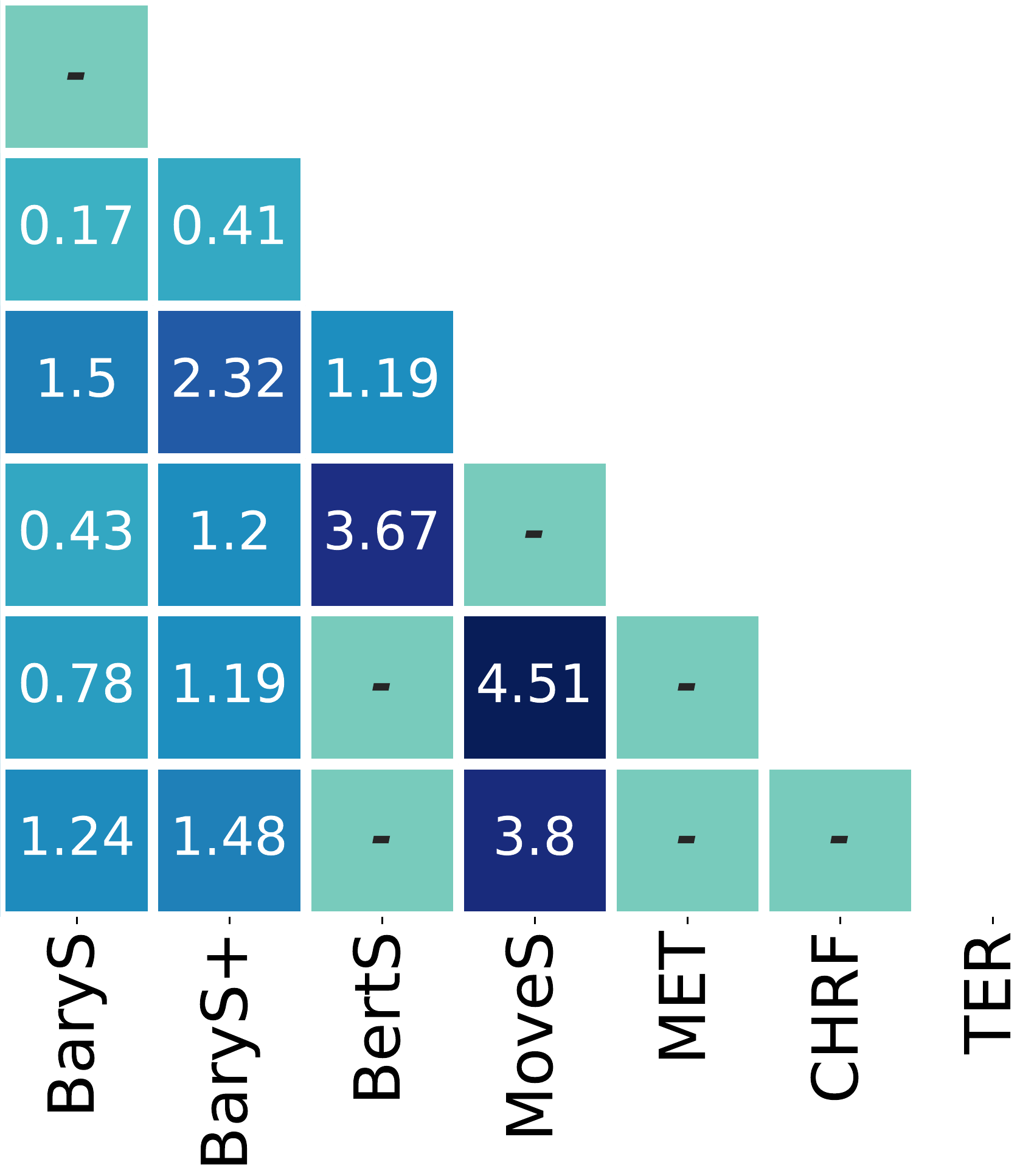} }
    \caption{Relevance}
    \end{minipage}
        \caption{$p$ values of the William significance test on data2text generation.  }\label{fig:pvalue_de_en}
\end{figure*}
We gather additional experimental results. In particular, a statistical analysis on WMT16 and WebNLG2020. 
\subsection{Statistical analysis of WMT16}
We report on \autoref{fig:kendall_matrix_de_en} the correlation coefficients the inter-correlation across metrics on WMT16.
\subsection{Statistical analysis of data2text}
We report in \autoref{fig:pvalue_de_en} the results of the William test on data2text generation.
\cite{quentin_lhoest_2021_5071218}
\end{document}